\documentclass{article}
\usepackage{iclr2027_conference,times}

\usepackage{amsmath,amsfonts,bm}

\def\eqref#1{equation~\ref{#1}}

\def\1{\bm{1}}

\DeclareMathAlphabet{\mathsfit}{\encodingdefault}{\sfdefault}{m}{sl}
\SetMathAlphabet{\mathsfit}{bold}{\encodingdefault}{\sfdefault}{bx}{n}

\usepackage{hyperref}
\usepackage{url}
\usepackage{graphicx}
\usepackage{booktabs}
\usepackage{tabularx}
\usepackage{multirow}
\usepackage{amsmath}
\usepackage{xcolor}
\usepackage{tikz}
\usepackage{pifont}
\usepackage{wrapfig}
\usepackage{longtable}
\usepackage{capt-of}
\usepackage{array}
\usepackage{placeins}
\usepackage{enumitem}
\usetikzlibrary{positioning,fit,arrows.meta,calc,backgrounds}
\usepackage{fontawesome5}

\newcommand{\ourBench}{{SRE-Marathon}}
\newcommand{\ourScore}{{Marathon-Score}}
\newcommand{\gptluna}{GPT\nobreakdash-5.6~Luna}

\input{results/macros.tex}
\input{results/macros_lists.tex}
\input{results/macros_marathon.tex}
\input{results/macros_rq.tex}
\input{results/macros_cases.tex}

\title{SRE-Marathon: \\ A Continuous, Change-Driven Benchmark for Autonomous Site Reliability Agents}

\author{%
  Yifang Tian\textsuperscript{1},
  Yingjian Bai\textsuperscript{1},
  Yifeng He\textsuperscript{1},
  Zichun Chong\textsuperscript{1}, \\
  \bfseries
  Yuanchen Gao\textsuperscript{2},
  Yiran Li\textsuperscript{1},
  Hans-Arno Jacobsen\textsuperscript{1} \\[4pt]
  \textsuperscript{1}University of Toronto \\
  \textsuperscript{2}The Hong Kong University of Science and Technology \\[4pt]
  \texttt{yifang.tian@mail.utoronto.ca, yingjian.bai@alumni.utoronto.ca} \\
  \texttt{\{yifeng.he, zichun.chong\}@mail.utoronto.ca, ygaocv@connect.ust.hk} \\
  \texttt{one.li@utoronto.ca, jacobsen@eecg.toronto.edu }
}

\iclrfinalcopy

\begin{document}

\maketitle

\lhead{Preprint}   

\begin{abstract}
Benchmarks for site reliability engineering (SRE) agents are typically episodic: one fault is injected, the agent receives an incident task, and its response is scored. Production operation is not. Incidents surface through noisy alerts, overlap in time, and often originate from code or configuration changes. We present SRE-Marathon, a benchmark for long-horizon, continuous SRE operation. An agent is invoked at a fixed cadence with cumulative alert history and a persistent workspace while operating a live two-zone Kubernetes deployment as a fault orchestrator injects overlapping faults according to a seeded, production-calibrated schedule. Curated code and configuration changes deployed through the same build pipeline available for repair. Each run is recorded into a sealed bundle and scored offline: Marathon-Score credits each injected fault for ordered progress through correlation, localization, and repair, with all metrics computed deterministically from recorded system evidence. Across three applications and about sixty faults per run, the best of \val{ms.methods.word} methods reaches only \val{ms.best.score} out of 100. Agents often correlate and localize faults, but almost never complete repairs while the faults remain active.

\end{abstract}

\section{Introduction}

Operating a live online service requires continuously keeping it available and reliable. When problems arise, an on-call engineer is notified through a stream of alerts, many of which may be noisy or redundant, and must determine whether an incident is occurring, which symptoms belong together, what changed, how to remedy it (roll back, reconfigure, or fix forward), and whether the service has recovered, all while failures degrade user experience and incur operational cost~\citep{zhang2025aiopssurvey}. Language-model agents are increasingly proposed for this role, and a growing line of benchmarks evaluates their ability to diagnose and mitigate failures~\citep{chen2025aiopslab,jha2025itbench,clark2026sregym}. Yet an important question remains unanswered: can these agents sustain production operations over time?

Production operations are not merely a sequence of isolated incident-resolution tasks. They involve continuously operating live services while problems are detected, disentangled, prioritized, and resolved. This setting is characterized by three essential features. \textit{First}, problems surface through noisy operational signals rather than explicit incident descriptions: an engineer observes an evolving alert stream and must determine whether an incident exists and which symptoms belong together. \textit{Second}, many incidents originate from changes within the system itself. A study of a large cloud service found that code and configuration bugs together account for 39.5\% of high-severity incidents~\citep{ghosh2022fight}, making diagnosis and repair of internal system changes central to production operations. \textit{Third}, incidents overlap and interventions have consequences. Multiple incidents may be active at once, while repairs consume time and resources and may disrupt otherwise unaffected parts of the system. Evaluation must therefore consider not only diagnostic accuracy, but also whether problems are resolved promptly, safely, and efficiently.

Existing benchmarks~\citep{jha2025itbench,clark2026sregym,gong2026orcabench,wang2026cloudopsbench} have largely overlooked these characteristics. They typically follow an episodic protocol: start from a clean environment, inject a single fault, assign the agent an explicit incident-resolution task, and terminate once it submits an answer. This setup provides the agent with an explicit incident context rather than evaluating whether it can detect, correlate, and triage noisy operational signals. Fault injection is typically based on chaos primitives, excluding failures caused by internal code or configuration changes and therefore removing rollback and fix-forward as repair options. Episodes typically contain one incident at a time and are evaluated independently, leaving overlap, accumulated mistakes, collateral effects, and operational cost largely unmeasured. On SREGym, frontier agents already achieve up to 72.6\% diagnosis success and 78.5\% mitigation success~\citep{clark2026sregym}. Such strong results are obtained under a substantially simplified model of production operations.

\begin{figure}[t]
\centering
\vspace{-3mm}
\includegraphics[width=\linewidth]{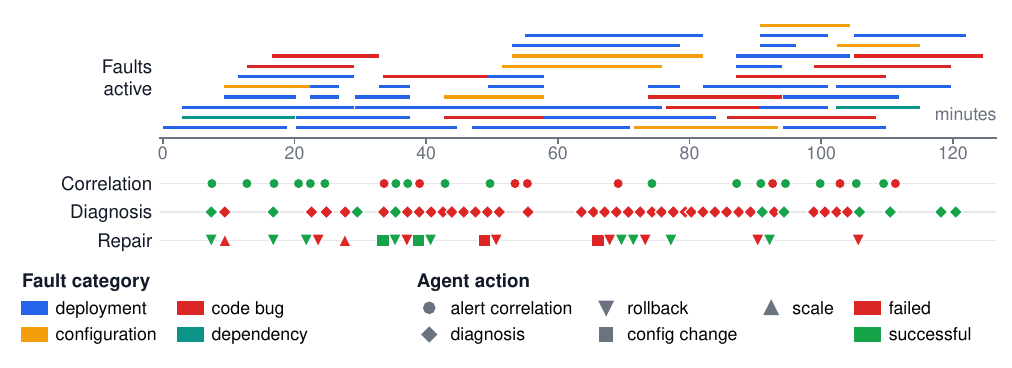}
\vspace{-10mm}
\caption{One \ourBench\ run using the ReAct~\citep{yao2023react} agent with Kimi K2.5 on Astronomy Shop. Above: every fault from occurrence to revert. Below: the agent's actions. The wall-clock axis runs longer than a nominal 90-minute run because the schedule drifts (Appendix~\ref{app:setup}).}
\vspace{-5mm}
\label{fig:motivation}
\end{figure}

We introduce {\ourBench}, a benchmark for sustained production operations rather than isolated incident episodes. The agent is invoked periodically to inspect the accumulated alert history and operate a live distributed microservice deployment under realistic workload. Unlike benchmarks that rely primarily on externally induced chaos faults, \ourBench\ also introduces failures through realistic code and configuration changes that pass basic validation before deployment, requiring the agent to reason about rollback, reconfiguration, and fix-forward. Faults are injected continuously according to seeded, production-calibrated schedules whose intensity increases over time, creating overlapping incidents and sustained operational pressure. Evaluation spans the full run and the full incident lifecycle, including detection and correlation, root-cause localization, repair, and recovery validation. Figure~\ref{fig:motivation} illustrates these characteristics in one run.

To ensure reproducibility despite the live environment, \ourBench\ records each run into a sealed bundle and computes all metrics deterministically offline from system evidence. Across \val{ms.methods.word} methods and two model backbones, agents correlate many faults and some methods locate a substantial fraction of their root causes, but successful repair remains rare under sustained operation. The best \ourScore{} reaches only \val{ms.best.score} out of 100, revealing a substantial gap between current agents and end-to-end autonomous production operation.
Our contributions are:

\begin{itemize}[leftmargin=*,itemsep=1pt]

\item \textbf{A continuous benchmark for production operations} (\S\ref{sec:environment}), with overlapping faults, evolving alert streams, live microservice deployments, and realistic code and configuration changes.

\item \textbf{An offline evaluation protocol and metric suite} (\S\ref{sec:evaluation}) covering correlation, localization, repair, intervention risk, service degradation, latency, and cost, all computed deterministically from recorded system evidence.

\item \textbf{A broad empirical evaluation and failure analysis} (\S\ref{sec:experimentalsetup}--\S\ref{sec:cases}) across \val{ms.methods.word} methods spanning runbooks, SRE agents, and coding agents on two model backbones.

\end{itemize}

\section{The SRE-Marathon Design}
\label{sec:environment}
\label{sec:formulation}
\label{sec:live-system}
\label{sec:problem-director}
\label{sec:observability}
\label{sec:tool-plane}
\label{sec:agent-plane}

\begin{figure*}[t!]
    \centering
    \centerline{
        \includegraphics[width=\linewidth]{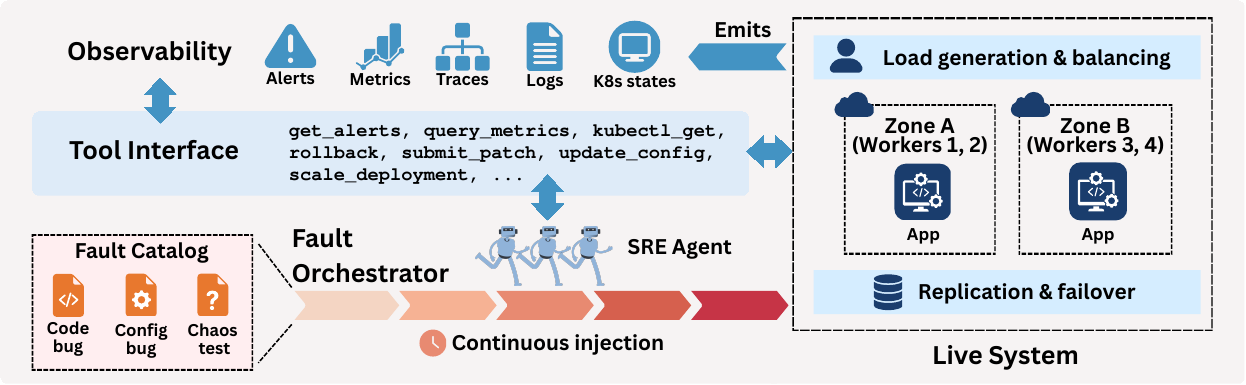}
    }
    \vspace{-3mm}
    \caption{Overview of \ourBench. A fault orchestrator continuously introduces problems into a live microservice system under load. The SRE agent observes their effects through telemetry and alerts and interacts with the system only through instrumented tools.}
    \label{fig:bench}
\end{figure*}

\ourBench\ is designed to evaluate SRE agents under sustained production operation. 
To do so, it places an agent in charge of a live microservice application while faults arrive continuously according to a seeded schedule. 
Failures may overlap, propagate across services, and compete for the agent's attention. 
Rather than receiving an incident-specific task, the agent is invoked periodically and must infer problems from operational signals, decide which incidents require attention, and determine when and how to intervene (Figure~\ref{fig:bench}).

\paragraph{Overview.}
A benchmark run consists of three components.
First, a \emph{live system} hosts a microservice application under continuous background traffic, so failures and repairs have observable effects on user-facing service behavior.
Second, a \emph{fault orchestrator} generates a reproducible sequence of code, configuration, deployment, dependency, infrastructure, and other failures from a seeded schedule.
Each injected problem has hidden ground truth, including its responsible service and artifact, that is retained by the benchmark but never exposed to the agent.
Third, an \emph{agent--system interface} exposes alerts, telemetry, system state, and controlled operational actions.
The agent is invoked periodically through this interface, with a persistent workspace across invocations, and every interaction is recorded.

\paragraph{Live system.}
\ourBench\ places the agent in a running microservice environment rather than replaying a fixed telemetry trace.
This is important because an intervention must be evaluated by what actually happens after it is executed: whether the affected service recovers, whether user-facing degradation decreases, and whether the action introduces collateral effects elsewhere in the system.

We use three microservice applications spanning different architectures, topologies, and failure surfaces: the 20-service OpenTelemetry Demo Astronomy Shop~\citep{oteldemo}, the 11-service Online Boutique~\citep{onlineboutique}, and the 41-service Train Ticket~\citep{zhou2018trainticket}.
Each run deploys one application under continuously generated workload.
For Astronomy Shop, we additionally emulate a geo-distributed deployment using two Kubernetes zones with replicated instances and failover, allowing individual components to fail without taking down the entire application and making cross-zone interventions visible in user-facing availability.

\paragraph{Fault Orchestrator.}
Long-horizon evaluation requires faults to arrive continuously while remaining reproducible across runs.
We therefore develop a seeded \emph{Fault Orchestrator} that determines which problems occur, where they occur, when they begin, and when reverted faults are removed.
Given the same benchmark release and random seed, the orchestrator is guaranteed to produce the same sequence of injected faults.

\textit{Fault catalog.}
The orchestrator draws from a catalog of \val{tax.total} faults across our three applications (Table~\ref{tab:taxonomy}).
Each catalog entry records its failure mechanism, target service, affected scope, symptom probe, and responsible artifact (Figure~\ref{fig:card}).
A fault is injected at a single target service, while any downstream symptoms arise through the application's own dependency structure.
This preserves an unambiguous injected root cause while allowing its effects to propagate across services and overlap with those of other faults.

\textit{Fault injection mechanisms.}
We support two mechanisms for introducing failures.
First, \emph{change-driven faults} are delivered as realistic code or configuration changes.
For code faults, a curated patch introduces a defect as a side effect of an otherwise plausible modification; the orchestrator deploys the change through the normal release pipeline.
Each patch passes the service's lint and unit tests, modeling a defect that escapes continuous integration.
Configuration faults are likewise introduced through changes to the running system.

\begin{wrapfigure}{r}{0.5\linewidth}
\vspace{-3mm}
\centering
\begin{minipage}{\linewidth}
\centering
\captionof{table}{Fault catalog across the three benchmark applications (details in Appendix~\ref{app:library}).}
\label{tab:taxonomy}
\vspace{1mm}
{\scriptsize\setlength{\tabcolsep}{5pt}

\begin{tabular}{@{}llrrrr@{}}
\toprule
Category & Mechanism & Astro & OB & TT & Total \\
\midrule
Code bug & change or direct   & 160 & 32 & 32 & 224 \\
Config bug & change or direct & 35  & 75 & 93 & 203 \\
Chaos test & direct & 122 & 66 & 85 & 273 \\
\midrule
\textit{Total} &  & 317 & 173 & 210 & 700 \\
\bottomrule
\end{tabular}}
\end{minipage}

\vspace{2mm}
\definecolor{cardCat}{RGB}{31,119,180}
\definecolor{cardPath}{RGB}{255,127,14}
\definecolor{cardTarget}{RGB}{44,160,44}
\definecolor{cardCone}{RGB}{148,103,189}
\definecolor{cardProbe}{RGB}{214,39,40}
\definecolor{cardArtifact}{RGB}{140,86,75}
\resizebox{\linewidth}{!}{%
\begin{tikzpicture}[
  line/.style={anchor=west, inner sep=1.2pt, font=\scriptsize\ttfamily},
  box/.style={draw, rounded corners=1.5pt, inner xsep=2pt, inner ysep=0.6pt, semithick},
  tag/.style={anchor=west, font=\scriptsize\rmfamily\bfseries, inner sep=1.5pt}
]
\foreach \i/\t in {
 0/{problem\_id: patch-upd-08-catalog-page-size},
 1/{category: code-bug \ \ family: code-change},
 2/{path: change\_engine},
 3/{services: [product-catalog]},
 4/{cone: [checkout, frontend, product-catalog,},
 5/{\ \ \ \ \ \ \ product-reviews, recommendation]},
 6/{symptom\_oracle\_id: probe-products},
 7/{probe: \{steps: [\{path: /api/products\}]\}},
 8/{culprit\_ref: patch:UPD-08},
 9/{artifact: \{kind: code\}},
 10/{schedulable: true}}
 \node[line] (l\i) at (0,-0.37*\i) {\t};
\begin{scope}[on background layer]
\node[draw=black!30, fill=black!3, rounded corners=2pt, inner sep=3pt, fit=(l0)(l10)(l4.east)] (card) {};
\end{scope}
\node[box, draw=cardCat,      fit=(l1)] (b1) {};
\node[box, draw=cardPath,     fit=(l2)] (b2) {};
\node[box, draw=cardTarget,   fit=(l3)] (b3) {};
\node[box, draw=cardCone,     fit=(l4)(l5)] (b4) {};
\node[box, draw=cardProbe,    fit=(l6)(l7)] (b5) {};
\node[box, draw=cardArtifact, fit=(l8)(l9)] (b6) {};
\node[tag, text=cardCat]      at (card.east |- b1) {category, failure mechanism};
\node[tag, text=cardPath]     at (card.east |- b2) {injection mechanism};
\node[tag, text=cardTarget]   at (card.east |- b3) {target service};
\node[tag, text=cardCone]     at (card.east |- b4) {affected scope};
\node[tag, text=cardProbe]    at (card.east |- b5) {symptom probe};
\node[tag, text=cardArtifact] at (card.east |- b6) {artifact};
\end{tikzpicture}}
\vspace{-5mm}
\caption{Example fault specification. The highlighted fields define its category, injection mechanism, target service and affected scope, symptom probe, and responsible artifact.}
\label{fig:card}
\vspace{-4mm}
\end{wrapfigure}

Second, \emph{direct faults} introduce failures through controlled interventions to the deployment, such as resource exhaustion, network disruption, or other chaos-style perturbations.
Every catalog entry is admitted only after injection reproduces the intended failure and the system successfully recovers afterward (Appendix~\ref{app:problem-generation}).

\textit{Marathon schedule.}
The orchestrator samples fault categories according to the production distribution reported by \citet{ghosh2022fight}: code bug \val{tax.code-bug.share}\%, dependency \val{tax.dependency.share}\%, infrastructure \val{tax.infrastructure.share}\%, deployment \val{tax.deployment.share}\%, configuration bug \val{tax.configuration.share}\%, database or network \val{tax.db-network.share}\%, and authentication \val{tax.auth.share}\%, renormalized over the categories supported by each application.
It then selects a compatible fault from the catalog.

In our main schedule, fault arrivals follow a seeded process whose intensity increases four-fold over the run, while individual faults remain active for four to twelve minutes before being reverted.
This increasing concurrency tests whether an agent can continue detecting, disentangling, and resolving incidents as operational load accumulates.
The agent receives no notification when a fault begins and can only detect it through its observable effects.

\paragraph{Agent--system interface.}
The agent interacts with the live system through \val{tools.published} instrumented tools exposed through the Model Context Protocol~\citep{mcp2024spec}, together with a periodic invocation protocol.
The tools provide access to operational telemetry and system state, support incident reporting, and execute controlled interventions.
All tool calls are recorded for offline evaluation (Appendix~\ref{app:tools}).

\textit{Observability tools.}
All three benchmark applications export the same classes of operational signals through OpenTelemetry Collectors~\citep{opentelemetry}: metrics are stored in Prometheus~\citep{prometheus}, traces in Jaeger~\citep{jaeger}, and logs in Loki~\citep{loki}.
We design a set of \val{alerts.total} alerting rules that continuously evaluates these signals and emits alerts when their conditions are met (Appendix~\ref{app:alerts}).
The resulting timestamped alerts form the alert history accessible to the agent via the \texttt{get\_alerts} tool.
Alerts are not grouped into incidents automatically, leaving the agent responsible for determining which alerts correspond to the same underlying problem.
We also make sure that telemetry emitted from the fault orchestrator and other benchmark infrastructure is excluded, so the agent observes only signals from the application and its deployment.

\textit{System interaction tools.}
In \ourBench, read-only tools expose alerts, metrics, logs, traces, service topology, Kubernetes state, and deployed source code.
The \texttt{incident\_report} tool lets the agent record incident grouping, root cause diagnosis, and repair status.
Operational tools support actions including restart, scaling, rollback, configuration changes, feature-flag updates, and source-level repair.
Most of those interventions are exposed through a corresponding tool; source repair follows a three-step workflow:
\texttt{propose\_patch} applies a source diff,
\texttt{build\_service\_image} builds the modified service, and
\texttt{submit\_patch} publishes and deploys it.

Because these tools can modify a live system, mutating operations pass through a policy enforcement layer that checks their targets, confines actions to benchmark resources, enforces per-invocation limits, and records forbidden operations.
This provides a controlled action boundary while preserving the effects of the agent's interventions on the running system.

\textit{Periodic invocation.}
We invoke the agent at a fixed interval $\Delta$ throughout the run.
At each interval, the agent receives the cumulative alert history together with its persistent private workspace.
A new invocation begins only after the previous one has finished, and each invocation runs in a fresh process while the workspace persists across intervals.
Compared to invoking agents on every alert, this design grants every agent a fixed number of opportunities to reason and act over a \ourBench\ run, regardless of the alert volume.

Each invocation is given a wall-clock budget shorter than $\Delta$; for the long runs in our evaluation, $\Delta=120$\,s and the invocation budget is 110\,s.
If the budget is exceeded, the invocation terminates while the live system and fault schedule continue.
Section~\ref{sec:experimentalsetup} reports the remaining run parameters.
\section{Evaluation}
\label{sec:evaluation}
\label{sec:bundle}
\label{sec:attribution}
\label{sec:families}
\label{sec:smash}

Evaluating a long-running SRE agent requires more than checking its final diagnosis.
We organize incident handling into three stages: (a)~\emph{correlation}, which associates alerts with incidents; (b)~\emph{localization}, which identifies the responsible service and artifact; and (c)~\emph{repair}, which verifies that the underlying fault has been corrected.
We report a metric for each stage and combine them in the \ourScore{} to measure end-to-end progress through the incident lifecycle.
Because all metrics are derived from recorded system evidence rather than a model judge, they may also serve as verifiable objectives for training SRE agents~\citep{stojanovski2026reasoning,pan2024swegym}.

Each run records the injected faults, alert history, telemetry, agent reports, tool calls, executed interventions, and user-facing service measurements, allowing all metrics to be recomputed offline.
We write $\mathcal{F}$ for the set of injected faults and $\mathcal{P}_f$ for the affected scope of fault $f$, i.e., the services that may be affected through the application's dependency graph.

\paragraph{The \ourScore.}
The \ourScore{} summarizes how successfully the agent progresses each injected fault through correlation, localization, and repair, while accounting for severe safety violations.
For each fault $f\in\mathcal{F}$, let $a_f$, $b_f$, and $c_f$ indicate successful correlation, localization, and repair, respectively.
Let $d=1$ if the run contains an attempted forbidden operation and $d=0$ otherwise:
\begin{equation}
\text{\ourScore} =
100\cdot
\Bigl(1-\frac{d}{2}\Bigr)
\cdot
\frac{1}{|\mathcal{F}|}
\sum_{f\in\mathcal{F}}
\frac{a_f + a_f b_f + a_f b_f c_f}{3}.
\label{eq:score}
\end{equation}
The three terms enforce the lifecycle order: localization contributes only after successful correlation, and repair only after both correlation and localization.
A fault that passes all three stages is \emph{handled} and receives full credit.
We report both the aggregate \ourScore{} and the individual stage metrics.
The denominator includes every injected fault, including faults whose symptoms disappear before the next measurement, preventing rapid repairs from reducing the number of faults being evaluated.
Appendix~\ref{app:worked} provides a worked example.

\begin{itemize}[leftmargin=1.6em,itemsep=4pt,topsep=3pt]

\item[(a)] \textbf{Correlated.}
Correlation measures whether the agent correctly associates a fault with an incident, with
$\mathrm{Corr}=\frac{1}{|\mathcal{F}|}\sum_f a_f$.
For each fault, we identify alerts consistent with its active time window, namespace, and affected scope $\mathcal{P}_f$, then match faults one-to-one with incidents reported by the agent.
We set $a_f=1$ when the matched incident contains at least one alert attributable to $f$.
Because overlapping faults may affect the same services, an alert can be compatible with more than one active fault; this occurs for \val{ms.ambiguous.OB}\%, \val{ms.ambiguous.TT}\%, and \val{ms.ambiguous.Astro}\% of alerts in the three applications.

\item[(b)] \textbf{Located.}
Localization measures whether the agent correctly identifies both where a fault occurred and what artifact caused it, with
$\mathrm{Loc}=\frac{1}{|\mathcal{F}|}\sum_f b_f$.
We set $b_f=1$ when the diagnosis associated with $f$ exactly identifies its culprit service or services and its responsible artifact.

\item[(c)] \textbf{Repaired.}
Repair measures whether an intervention by the agent resolves the fault while it is still active, with
$\mathrm{Rep}=\frac{1}{|\mathcal{F}|}\sum_f c_f$.
We set $c_f=1$ when the agent executes an intervention targeting a service in $\mathcal{P}_f$ and subsequent system evidence confirms recovery.
Recovery is established when the fault's symptom check returns to and remains in a healthy state, or, when no symptom failure was observed, when the responsible artifact is verified to have been corrected.
Only interventions completed before the orchestrator reverts the fault are eligible, so benchmark cleanup is never credited as an agent repair.

\item[(d)] \textbf{Intervention risk.}
Intervention risk measures the fraction of actions that either extend beyond the affected scope of the referenced incident or violate an explicit safety policy:
$\mathrm{Risk}=(n_{\mathrm{out}}+n_{\mathrm{forb}})/(n_{\mathrm{exec}}+n_{\mathrm{forb}})$.
Here, $n_{\mathrm{exec}}$ is the number of executed interventions, $n_{\mathrm{out}}$ counts actions outside the affected scope, and $n_{\mathrm{forb}}$ counts attempted forbidden operations such as deleting the benchmark namespace or accessing protected secrets.

\end{itemize}

\paragraph{Operational outcomes.}
Beyond incident-handling effectiveness, we report service impact, response latency, and computational cost.

\begin{itemize}[leftmargin=1.6em,itemsep=4pt,topsep=3pt]

\item[(e)] \textbf{Service degradation.}
Service degradation measures the cumulative amount of user-facing traffic affected by errors or excessive latency.
At time $t$, let $e(t)$ be the failed-request fraction, $\ell(t)$ the fraction of requests slower than 2.5\,s, and $q(t)$ the request rate at the application's entry-point service.
We define service degradation $D=\int \min\{1,e(t)+\ell(t)\}\,q(t)\,dt$, where the first term estimates the fraction of degraded requests and $q(t)$ weights it by traffic volume.
We report the degradation level, $\mathrm{Degrad.\ Level}=D_{\mathrm{agent}}/D_{\mathrm{no\text{-}action}}$, as a percentage, against a no-action run with the same application configuration and fault-schedule seed.

\item[(f)] \textbf{Time to detect, locate, and repair.}
For each fault, MTTD is the elapsed time from injection to the first incident report containing one of its alerts, MTTL to the first correct localization, and MTTR to the first verified repair.
We report the median for each metric over faults that successfully reach the corresponding stage.
Because these metrics are computed over different subsets of faults, their values need not be ordered; for example, MTTL can be lower than MTTD.
All timing measurements have resolution $\Delta$, the agent invocation interval.

\item[(g)] \textbf{Cost.}
We report the total model calls, token consumption, and estimated monetary cost of each run.
Tokens are the primary model-independent usage measure, while dollar cost reflects the pricing used in our experiments.

\end{itemize}
\section{Experimental setup}
\label{sec:experimentalsetup}

\paragraph{Schedules and deployment environments.}
Our main evaluation follows the \emph{marathon} schedule defined in \S\ref{sec:problem-director}.
Each run injects around sixty faults over a 60-minute period of increasing intensity, followed by a 30-minute observation period with no new injections.
The agent is invoked every $\Delta=120$\,s.
Across Online Boutique, Train Ticket, and Astronomy Shop, this schedule results in an average of \val{ms.live.mean.OB}, \val{ms.live.mean.TT}, and \val{ms.live.mean.Astro} simultaneously active faults, respectively, with peak concurrency of \val{ms.live.peak.OB}, \val{ms.live.peak.TT}, and \val{ms.live.peak.Astro}.
The schedule is intentionally designed to stress sustained operation under increasing fault concurrency rather than to reproduce the incident rate of a typical production hour.

For Marathon-Score evaluation, all methods share the same seeded schedule for each of the three applications.
The main experiments run on a single-host Kind cluster~\citep{kind}, which provides a reproducible environment for controlled comparison.
We additionally validate the benchmark on a seven-node kubeadm deployment (Appendix~\ref{app:setup}).

\paragraph{Methods.}
We compare ten methods. Five are baselines implemented by us (Appendix~\ref{sec:orbit}): a non-LLM rule-based \emph{Runbook}; a tool-augmented \emph{ReAct} agent~\citep{yao2023react}; \emph{RAG}~\citep{lewis2020rag}, which retrieves live telemetry and produces one diagnosis-and-repair plan per invocation; \textsc{Orbit}, which maintains persistent incident state across invocations; and \emph{Keep-agent}, which uses Keep's~\citep{keephq2026keep} rule engine for alert grouping and one LLM call per invocation for diagnosis and repair planning. We additionally evaluate the SRE agent \emph{OpenSRE}~\citep{tracer2026opensre}, the coding agents \emph{Codex}~\citep{openai2026codex} and \emph{Claude Code}~\citep{anthropic2026claudecode}, both unmodified (Appendix~\ref{sec:orbit}), and the diagnosis methods \emph{OpsMem}~\citep{sun2026opsmem} and \emph{COCA}~\citep{li2025coca}, which execute interventions through our minimal repair shell. All methods share the same alerts, tools, and policy enforcement layer. We evaluate the LLM-based methods with Kimi K2.5~\citep{moonshot2025kimi} and \gptluna~\citep{openai2026gpt56}, under a budget of less than \$8 per run.

\section{Results}
\label{sec:results}

\begin{table}[t!]
\vspace{-3mm}
\caption{\ourBench\ leaderboard: each row averages a method's marathon runs on three microservice applications; the best value per column is in bold.}
\label{tab:pooled}
\begin{center}
\scriptsize
\setlength{\tabcolsep}{3.2pt}
\resizebox{\linewidth}{!}{\begin{tabular}{@{}llrrrrrrrrrrc@{}}
\toprule
 & & \multicolumn{3}{c}{Stage metrics} & \multicolumn{3}{c}{Median time (s)} & \multicolumn{1}{c}{\multirow{2}{*}{\shortstack{Degrad.\\Level (\%)}}} & \multicolumn{2}{c}{Cost per run} & \multicolumn{1}{c}{\multirow{2}{*}{Risk}} & \multirow{2}{*}{\shortstack{\ourScore\\(Out of 100)}} \\
\cmidrule(lr){3-5}\cmidrule(lr){6-8}\cmidrule(lr){10-11}
Method & \multicolumn{1}{c}{Backbone} & \multicolumn{1}{c}{Corr} & \multicolumn{1}{c}{Loc} & \multicolumn{1}{c}{Rep} & \multicolumn{1}{c}{MTTD} & \multicolumn{1}{c}{MTTL} & \multicolumn{1}{c}{MTTR} &  & \multicolumn{1}{c}{Tokens} & \multicolumn{1}{c}{\$} &  &  \\
\midrule
Runbook & --- & 0.658 & 0.364 & \textbf{0.024} & 167 & 267 & 560 & 111.0 & --- & --- & 0.028 & 33.5 \\
\midrule
ReAct & kimi-k2.5 & 0.694 & 0.079 & 0.006 & 189 & 283 & 709 & 86.9 & 6.5M & 4.02 & 0.217 & 25.4 \\
\citep{yao2023react} & gpt-5.6-luna & 0.419 & 0.125 & 0.000 & 163 & 293 & --- & 100.8 & 6.3M & 1.45 & 0.349 & 16.3 \\
\specialrule{\cmidrulewidth}{0.2ex}{0.5ex}
RAG & kimi-k2.5 & 0.426 & 0.398 & 0.000 & 200 & 68 & --- & 150.5 & 569k & 0.46 & 0.015 & 23.5 \\
\citep{lewis2020rag} & gpt-5.6-luna & 0.199 & 0.105 & 0.000 & 159 & \textbf{65} & --- & 93.5 & 730k & 0.23 & \textbf{0.000} & 8.2 \\
\specialrule{\cmidrulewidth}{0.2ex}{0.5ex}
\textsc{Orbit} & kimi-k2.5 & 0.768 & 0.000 & 0.000 & 175 & --- & --- & 92.8 & 5.4M & 3.35 & --- & 25.6 \\
 & gpt-5.6-luna & 0.768 & 0.009 & 0.000 & 186 & 249 & --- & 98.6 & 7.1M & 1.63 & --- & 25.9 \\
\specialrule{\cmidrulewidth}{0.2ex}{0.5ex}
OpenSRE & kimi-k2.5 & 0.093 & 0.200 & 0.000 & 297 & 188 & --- & 99.7 & 7.1M & 4.40 & \textbf{0.000} & 3.7 \\
\citep{tracer2026opensre} & gpt-5.6-luna & 0.222 & 0.034 & 0.000 & \textbf{143} & 173 & --- & 92.5 & 7.0M & 1.59 & 0.158 & 8.3 \\
\specialrule{\cmidrulewidth}{0.2ex}{0.5ex}
Keep-agent & kimi-k2.5 & \textbf{0.775} & \textbf{0.491} & 0.017 & 188 & 165 & 476 & 92.1 & 685k & 0.72 & 0.014 & \textbf{41.3} \\
\citep{keephq2026keep} & gpt-5.6-luna & 0.656 & 0.313 & 0.017 & 184 & 142 & 455 & 84.5 & 742k & 0.31 & \textbf{0.000} & 31.3 \\
\specialrule{\cmidrulewidth}{0.2ex}{0.5ex}
Codex & kimi-k2.5 & 0.231 & 0.039 & 0.011 & 210 & 335 & 456 & 131.7 & 7.2M & 4.45 & 0.171 & 8.1 \\
\citep{openai2026codex} & gpt-5.6-luna & 0.502 & 0.138 & 0.023 & 203 & 346 & \textbf{411} & 85.6 & 8.0M & 1.81 & 0.104 & 20.4 \\
\specialrule{\cmidrulewidth}{0.2ex}{0.5ex}
Claude Code & kimi-k2.5 & 0.379 & 0.050 & 0.000 & 337 & 210 & --- & 103.1 & 6.7M & 4.14 & \textbf{0.000} & 13.4 \\
\citep{anthropic2026claudecode} & gpt-5.6-luna & 0.407 & 0.070 & 0.000 & 360 & 684 & --- & 80.7 & 3.4M & 0.84 & \textbf{0.000} & 14.3 \\
\specialrule{\cmidrulewidth}{0.2ex}{0.5ex}
OpsMem & kimi-k2.5 & 0.587 & 0.006 & 0.000 & 146 & 351 & --- & 124.3 & 754k & 0.55 & 0.200 & 19.8 \\
\citep{sun2026opsmem} & gpt-5.6-luna & 0.613 & 0.034 & 0.000 & 157 & 450 & --- & \textbf{77.4} & 984k & 0.25 & \textbf{0.000} & 21.0 \\
\specialrule{\cmidrulewidth}{0.2ex}{0.5ex}
COCA & kimi-k2.5 & 0.576 & 0.017 & 0.023 & 168 & 336 & 786 & 132.3 & 511k & 0.36 & 0.051 & 19.6 \\
\citep{li2025coca} & gpt-5.6-luna & 0.634 & 0.253 & 0.011 & 161 & 222 & 461 & 87.6 & \textbf{403k} & \textbf{0.09} & 0.103 & 28.0 \\
\bottomrule
\end{tabular}
}
\end{center}
\end{table}

\subsection{Overall benchmark performance}
\label{sec:results-overall}

\paragraph{End-to-end resolution remains rare, with repair as the primary bottleneck.}
Keep-agent achieves the highest \ourScore{} in Table~\ref{tab:pooled} at \val{ms.7.keep-llm.kimi.score}, followed by Runbook at \val{ms.7.b1-runbook.none.score}; the remaining Kimi K2.5 methods score between \val{ms.7.opensre-x3.kimi.score} and \val{ms.7.orbit-0.kimi.score}, while \gptluna{} methods range from \val{ms.7.rag-b2r.luna.score} to \val{ms.7.keep-llm.luna.score}. Despite these nonzero scores, complete resolution is extremely uncommon: only Keep-agent handles any fault end to end (\val{ms.7.keep-llm.kimi.OB.count} on Online Boutique with Kimi K2.5 and \val{ms.7.keep-llm.luna.OB.count} with \gptluna). Most score therefore comes from correlation and localization rather than repair; for example, \textsc{Orbit} obtains \val{ms.7.orbit-0.kimi.score} from correlation alone. Localization is strongest for Keep-agent (\val{ms.7.keep-llm.kimi.located}), RAG (\val{ms.7.rag-b2r.kimi.located}), and Runbook (\val{ms.7.b1-runbook.none.located}), whereas no coding or diagnosis method exceeds these baselines. Across all methods, repair remains at most \val{ms.7.b1-runbook.none.repaired} while faults are active. The strongest methods therefore rely heavily on explicit rule-based structure, while more fully agentic approaches still struggle to convert diagnosis into successful intervention.

\paragraph{Detection is relatively fast, but user impact and efficiency vary substantially across methods and backbones.}
On Kimi K2.5, methods that correlate faults at scale detect them within \val{ms.7.opsmem.kimi.time_to_detect}--\val{ms.7.react-b2.kimi.time_to_detect}\,s, corresponding to roughly one to two agent invocations. User-facing degradation, however, is much less consistent. RAG, Codex, OpsMem, and COCA worsen degradation on Kimi K2.5 (degradation level \val{ms.7.opsmem.kimi.level}\% to \val{ms.7.rag-b2r.kimi.level}\%) but improve it on \gptluna{} (\val{ms.7.opsmem.luna.level}\% to \val{ms.7.rag-b2r.luna.level}\%), while ReAct shows the opposite trend (\val{ms.7.react-b2.kimi.level}\% versus \val{ms.7.react-b2.luna.level}\%). This backbone sensitivity suggests that successful diagnosis alone is not sufficient to predict user-facing benefit. Cost also shows little correspondence with outcome: on Kimi K2.5, RAG and Keep-agent use \val{ms.7.rag-b2r.kimi.tokens} and \val{ms.7.keep-llm.kimi.tokens} tokens per run, compared with \val{ms.7.orbit-0.kimi.tokens}--\val{ms.7.codex-x4.kimi.tokens} for the more tool-intensive agents, without a consistent performance advantage. No method attempts a forbidden operation, so Risk reflects only out-of-scope actions; ReAct is highest at \val{ms.7.react-b2.kimi.risk} on Kimi K2.5 and \val{ms.7.react-b2.luna.risk} on \gptluna.

\paragraph{The repair bottleneck persists across backbones and seeds, while lower load improves end-to-end success.}
On \gptluna, localization reaches at most \val{ms.7.keep-llm.luna.located} and repair at most \val{ms.7.codex-x4.luna.repaired}, showing that the same failure mode is not specific to Kimi K2.5. The conclusion also holds across two additional Astronomy Shop seeds, where ReAct scores \val{ms.17.react-b2.kimi.score} and \val{ms.27.react-b2.kimi.score} and repairs at most \val{ms.17.react-b2.kimi.repaired} of faults (Appendix~\ref{app:robustness}, Table~\ref{tab:react-seeds}). Under lower load, however, the same ReAct agent scores \val{ms.held5.react.score} on five-fault runs and \val{ms.held10.react.score} on ten-fault runs, compared with \val{ms.storm.react.score} on marathon runs, and handles \val{ms.held5.react.handled}\%, \val{ms.held10.react.handled}\%, and \val{ms.storm.react.handled}\% of faults end to end (Appendix~\ref{app:short}, Table~\ref{tab:storm-drill}). Because the short runs also differ in venue, invocation interval, and seed, this comparison does not isolate load experimentally, but it is consistent with increasing concurrency being a major source of failure.

\begin{figure}[t]
\centering
\includegraphics[width=\linewidth]{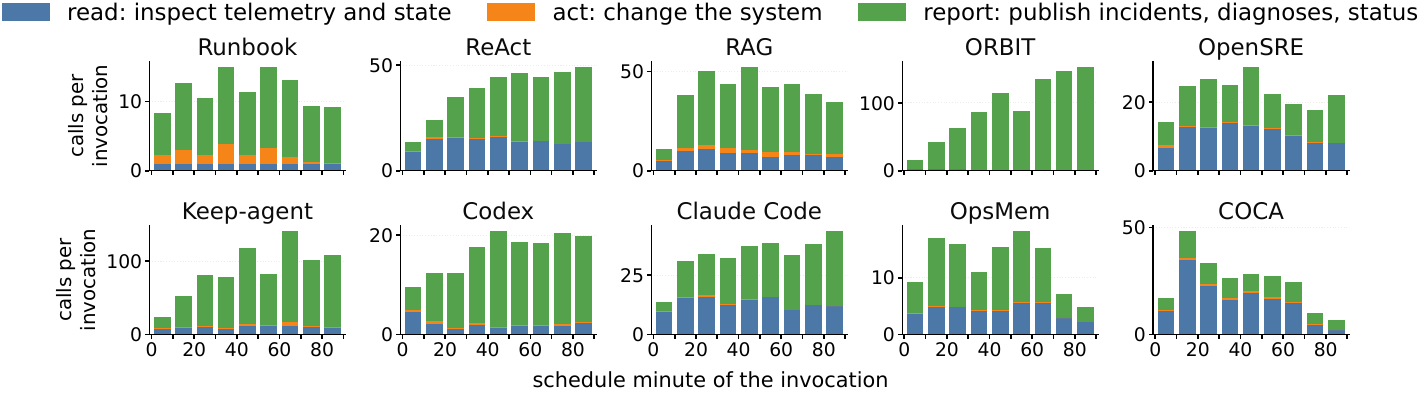}
\vspace{-6mm}
\caption{
Agent actions over time during a marathon run: tool calls per invocation, by tier, in ten-minute bins of the
schedule, for every method of Table~\ref{tab:pooled} on Kimi K2.5 and for Runbook.
}
\label{fig:activity-time}
\end{figure}

\begin{figure}[t]
\centering
\includegraphics[width=\linewidth]{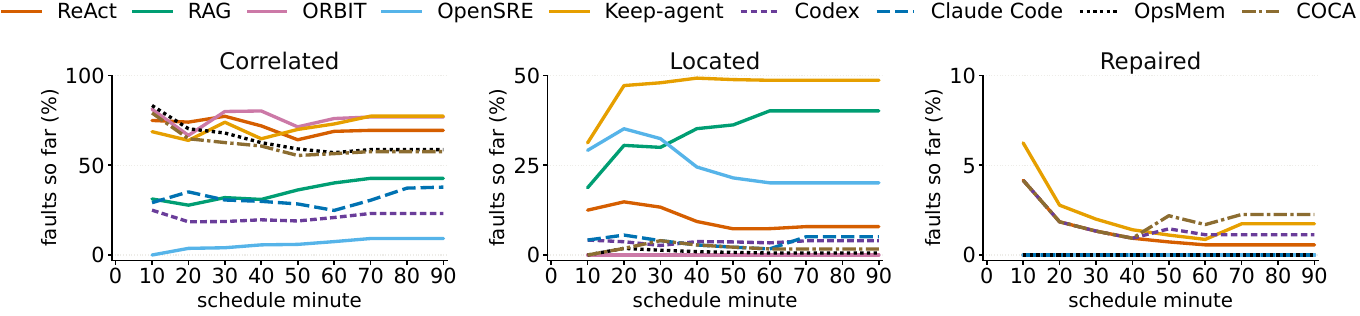}
\vspace{-6mm}
\caption{
Stage-wise success over time in fault handling. Each panel shows the \% of faults injected so far that have been successfully correlated, located, or repaired by each method on Kimi K2.5.
}
\label{fig:pipeline-time}
\end{figure}

\subsection{Performance analysis under sustained operation}
\label{sec:results-time}

Figures~\ref{fig:activity-time} and~\ref{fig:pipeline-time} show how agent activity and stage-wise success evolve over a marathon run, using the Kimi K2.5 runs from Table~\ref{tab:pooled}.

\paragraph{As fault load increases, agents report more without acting more.}
Figure~\ref{fig:activity-time} groups tool calls into reads, system-changing actions, and incident reports.
Call volume rises during the injection period for most LLM-based methods, but the increase is dominated by reporting.
ReAct's report calls grow from \val{rq.react-b2.kimi.0-10.claim} to \val{rq.react-b2.kimi.80-90.claim} per invocation, while its reads change only from \val{rq.react-b2.kimi.0-10.read} to \val{rq.react-b2.kimi.80-90.read}; \textsc{Orbit} shows a similar increase from \val{rq.orbit-0.kimi.0-10.claim} to \val{rq.orbit-0.kimi.80-90.claim}.
Codex's reads decline as its reporting increases.
Actions remain sparse, never exceeding \val{rq.act.max} per invocation in any bin, and by the final ten minutes only RAG and Keep-agent still average at least one action per invocation.
Thus, increasing load produces substantially more incident reporting, but little additional intervention.

\paragraph{Correlation remains stable, localization separates methods, and repair remains rare.}
Figure~\ref{fig:pipeline-time} reports the fraction of faults injected so far that have been successfully correlated, located, or repaired.
Correlation remains relatively stable as faults accumulate, with several methods maintaining similar success rates throughout the run.
Localization shows greater separation: Keep-agent and RAG improve or remain strong, while OpenSRE declines from \val{rq.opensre-x3.kimi.by10.located} to \val{rq.opensre-x3.kimi.by60.located} and most other methods remain low.
Repair is consistently the weakest stage; no Kimi K2.5 method exceeds \val{rq.keep-llm.kimi.by10.repaired} of injected faults at any point, and the few methods with early repairs generally do not sustain that rate as the run progresses.
Overall, agents continue to correlate incidents under increasing load, but only a few translate this into reliable localization, and almost none into successful repair.

\subsection{Case studies}
\label{sec:cases}

\begin{figure}[t]
\centering
\vspace{-2mm}
\includegraphics[width=\linewidth]{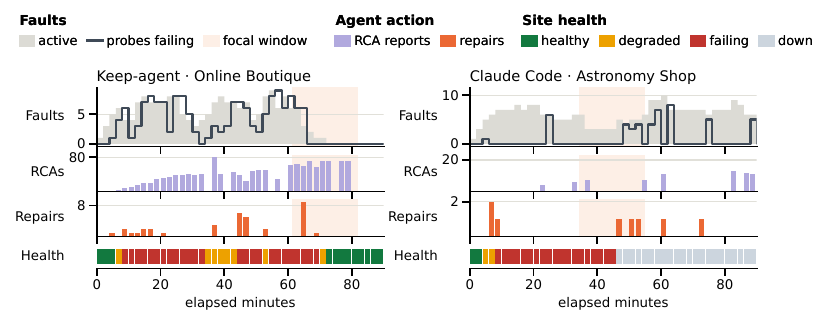}
\vspace{-10mm}
\caption{Two marathon runs containing the same fault type and correct remedy. The shaded focal window marks the lifetime of the fault discussed in each case. From top to bottom, the lanes show active faults (grey) and those with a failing symptom probe (black), RCA (Root Cause Analysis) reports, executed repairs, and user-facing service health.}
\label{fig:cases}
\end{figure}

Figure~\ref{fig:cases} compares two runs containing the same fault type, a bad image tag whose correct remedy is rollback. The shaded focal window marks the lifetime of the fault discussed below.

On \val{case.act_ok.app}, the fault is injected on \texttt{\val{case.act_ok.service}} at minute \val{case.act_ok.onset_min} (schedule drift; Appendix~\ref{app:setup}), while \val{case.act_ok.concurrent} other faults are active. Keep-agent groups the relevant alerts, identifies the service and artifact, and rolls the service back at minute \val{case.act_ok.acted_min}. The probe recovers by minute \val{case.act_ok.repaired_min} and remains healthy until the fault is reverted at minute \val{case.act_ok.revert_min}, so the fault is handled end to end.

On \val{case.act_no.app}, Claude Code also identifies the affected service, but first attempts two ineffective configuration changes and a manifest patch rejected by the policy enforcement layer. It reaches the correct rollback only at minute \val{case.act_no.rollback_min}, after \val{case.act_no.calls_before_rollback} attempts, too late for the repair to take effect before the fault is reverted at minute \val{case.act_no.revert_min}. This illustrates a central failure mode: agents can identify the right cause yet still fail to remediate it because of inefficient action selection and delayed intervention. Additional failure analysis appears in Appendix~\ref{app:decomposition}.

\section{Related Work}
\label{sec:related}

\paragraph{Benchmarks for agents.}
SRE benchmarks either evaluate diagnosis over frozen telemetry~\citep{xu2025openrca,pham2025rcaeval,fang2026openrca2}, expose a fixed incident through mocked tools or a recorded window~\citep{wang2026cloudopsbench,gong2026orcabench}, or allow an agent to observe and modify a live cluster one fault episode at a time~\citep{chen2025aiopslab,jha2025itbench,clark2026sregym}. Beyond SRE, agent benchmarks evaluate closed-loop behavior against tests, policies, and cost~\citep{jimenez2024swebench,yao2025taubench,merrill2026terminalbench}, assess coherence over long horizons~\citep{backlund2025vendingbench,xu2024theagentcompany}, and study failures caused by reward design and non-reproducible scoring~\citep{zhu2025abc}. \ourBench\ extends long-horizon evaluation to SRE through continuous operation on a live system, overlapping faults including defects delivered through the build pipeline, and deterministic scoring from sealed run recordings (\S\ref{sec:evaluation}).

\paragraph{LLM agents for the incident lifecycle.}
Most SRE agents target a particular stage of the incident lifecycle: alert correlation and rule generation~\citep{kuang2024cola,jiang2025logpilot,gu2025argos,wang2024comet}, root-cause analysis from tickets, telemetry, and code~\citep{chen2024rcacopilot,jiang2024xpert,wang2024rcagent,roy2024exploring,pei2025flowofaction,tian2026gala}, or mitigation from runbooks~\citep{an2024nissist,mao2026stepfly,zhang2024flash}; multi-agent pipelines span detection through repair~\citep{chen2025stratus,luo2026opsagent}. Deployed systems include coding agents applied to clusters~\citep{anthropic2026claudecode,openai2026codex}, open-source SRE agents and alert platforms~\citep{tracer2026opensre,keephq2026keep}, and provider offerings~\citep{aws2025devopsagent,azure2025sreagent,resolveai2026,bharadwaj2026adobe}. 


\section{Conclusion}

\ourBench\ evaluates SRE agents under sustained production operation, where faults arrive through noisy alerts, overlap, and may arise from code or configuration changes. Its metrics are computed deterministically from recorded system evidence, enabling reproducible offline evaluation. Across \val{ms.methods.word} methods and two backbones, the best \ourScore{} reaches only \val{ms.best.score} out of 100. Agents often correlate and localize faults, but rarely convert diagnosis into timely repair as load increases. Appendix~\ref{sec:limitations} discusses the limitations.




\bibliography{references}
\bibliographystyle{iclr2027_conference}

\appendix

\section{Environment setup}
\label{app:setup}

Both venues of \S\ref{sec:experimentalsetup} run on the same OpenStack cloud and carry the same application, telemetry, and fault-injection stack, deployed from Helm charts pinned by archive hash and images pinned by digest.

\paragraph{Single-host venue.} One virtual machine with 16 vCPUs, 32\,GB of RAM, and a 300\,GB volume hosts a Kind cluster of one control-plane node and four workers, two per zone through the standard zone label. Latency of 25\,ms one way, about 50\,ms round trip, is injected between the zones with \texttt{tc netem}, which is typical of two sites on one continent. Calico replaces Kind's default network plugin so that network faults behave as they would on a real network, and etcd keeps its data in memory because the virtual disk's write latency destabilized the control plane. Storage is the node's local path, images are preloaded, and the cluster never pulls one during a run.

\paragraph{Multi-node venue.} Seven virtual machines form a kubeadm cluster: three control-plane nodes with 4 vCPUs and 8\,GB each, placed on distinct hypervisors behind a virtual address that survives failover, and four workers with 16 vCPUs, 32\,GB, and a 100\,GB volume each, two per zone. No latency is injected between the zones. An eighth machine serves as bastion, image build host, and the registry the cluster pulls images from, and persistent volumes are Ceph-backed. Automatic upgrades are disabled so that no unattended restart falls inside a run.

\paragraph{Applications.} Astronomy Shop is the OpenTelemetry Demo (Helm chart 0.40.10) with 20 services, built from a fork that carries the seeded defects and deployed as an active and a standby site across the two zones. Online Boutique is upstream v0.10.6, 11 services rebuilt with OpenTelemetry instrumentation. Train Ticket is the FudanSELab release at commit \texttt{313886e9}, 41 workloads on upstream images. Direct faults are injected with chaos engineering from a namespace the agent cannot see, and models are hosted services called through the metering proxy, so no host needs a GPU.

\paragraph{Schedule clock.} The fault schedule advances by invocation rather than by wall-clock time: faults are injected and reverted between invocations, and the next invocation starts only once they are in place. A change-driven fault is deployed and reverted through an image rollout that can take minutes, so runs with many such faults, notably on Astronomy Shop, drift beyond their nominal 90 minutes of wall-clock time, while every method still receives the same invocations and faults in the same order.

\section{The fault catalog}
\label{app:library}

Table~\ref{tab:catalog} lists the \val{tax.total} catalog faults by category and application. The seven categories are those of the incident study the schedule follows (\S\ref{sec:problem-director}). Change-driven code and configuration faults also carry one of the eight defect subtypes of \citet{ghosh2022fight}, plus latent defect enabled by a feature flag in code bug and deployed setting in config bug. A fault is categorized by the cause of its failure rather than by its injection mechanism. A patch whose only defect is a wrong setting is therefore a configuration fault even though it arrives as source code.
\begin{table}[!htbp]
\caption{Catalog faults by category, application, and defect subtype, with an example of each.}
\label{tab:catalog}
\begin{center}
\scriptsize
\setlength{\tabcolsep}{3pt}
\begin{tabular}{@{}lrrrr>{\raggedright\arraybackslash}p{0.5\linewidth}@{}}
\toprule
Category and defect subtype & Astro & OB & TT & Total & Example \\
\midrule
Code bug &  &  &  &  &  \\
\quad Feature or logic error & 48 & 6 & 6 & 60 & A new currency lookup omits a supported currency, so requests in that currency receive an incorrect default value. \\
\quad Incorrect constant or flag & 5 & 6 & 6 & 17 & A new length limit on cart identifiers is set too low, so the cart rejects identifiers it previously accepted. \\
\quad Component incompatibility & 23 & 6 & 6 & 35 & New code reads database rows by column name while the adapter still returns values by position, so the queries fail. \\
\quad Type, validation, or exception error & 29 & 6 & 6 & 41 & A new price check reads only the whole-number part of a price, so valid prices below one dollar are rejected. \\
\quad Backward incompatibility & 45 & 8 & 8 & 61 & An update changes the request fields a service expects without accepting the older format, so callers still sending it are rejected. \\
\quad Latent defect enabled by a feature flag & 10 & -- & -- & 10 & A memory leak or a failing checkout path switched on at runtime. \\
\midrule
Config bug &  &  &  &  &  \\
\quad Invalid setting & 1 & 6 & 6 & 13 & A dependency address is edited to a port the destination does not listen on, so calls to it are refused. \\
\quad Uncoordinated change & 5 & 6 & 6 & 17 & Traffic is forwarded to a new port while the destination still listens on the original one, so requests fail to connect. \\
\quad Inconsistent names or values & 7 & 6 & 6 & 19 & Code reads a setting under a new name while the deployment still supplies the old one, so the service runs without the value. \\
\quad Deployed setting & 22 & 57 & 75 & 154 & A readiness probe that cannot pass, a memory limit too small, a wrong DNS policy. \\
\midrule
Chaos test &  &  &  &  &  \\
\quad Deployment error & 67 & 32 & 44 & 143 & A bad image tag, a scale to zero, a missing toleration. \\
\quad Dependency failure & 20 & 22 & 23 & 65 & A downstream service made unreachable, slow, or lossy. \\
\quad Infrastructure & 9 & 1 & 3 & 13 & CPU, file-descriptor, or port exhaustion. \\
\quad Database / network & 10 & -- & -- & 10 & A DNS name that stops resolving, a network policy that isolates a service, a cache failure. \\
\quad Auth failure & 16 & 11 & 15 & 42 & An RBAC denial for an init container, a disrupted shared-cache credential. \\
\midrule
\textit{Total} & 317 & 173 & 210 & 700 &  \\
\bottomrule
\end{tabular}

\end{center}
\end{table}

\section{Generating code and configuration faults}
\label{app:problem-generation}

The change-driven faults of Table~\ref{tab:catalog} are the code bug and config bug rows with one of the eight defect subtypes. They are written by hand and validated on a live deployment as follows.

\paragraph{Authoring.} A code fault starts from a service's source. We design a feature addition or refactoring that carries an incorrect assumption about existing data, callers, or dependencies, and confirm that ordinary requests reach the defective code. A change that fails only for a benchmark-specific input is rejected. A configuration fault changes a setting written directly into a deployment manifest: a dependency port, a resource limit, a connection host or password, a port on which callers and listener disagree, or a value cleared to model a lost setting. Indirectly loaded configuration, secrets, telemetry settings, and the applications' built-in fault controls are left alone. Each fault records the intended update, the defect, the requests that expose it, the log message it produces and the service that emits it, the patch or the changed values, and a recovery procedure. Agents see none of these records.

\paragraph{Validation.} Each fault is injected on the tested application version under the test workload, and two five-minute periods are compared, before and after the change. The recorded message must be absent before and present after, in the changed service or a direct dependant, and the application must keep serving requests. The fault must also meet one screening criterion: server errors averaging at least four per minute, at least ten new error log events per minute attributable to the change, or a service-instance restart. These thresholds screen for observable faults and do not measure degradation, since one failed request can produce several log events. For a code patch, text introduced by the patch must appear in the new image and in no earlier one, because a runtime error alone does not show that the patched image is running. For a configuration change the deployed value is checked, the change must still allow startup and readiness, and the whole test is repeated after the setting is restored, with only the second observation counting as evidence~\citep{yin2011configuration}. Finally the orchestrator's own injection and recovery are exercised. Recovery must restore the original image or setting, readiness, request processing, and a cleared message, and the other site must be unaffected. A fault validated this way, alone and within five minutes, may still be masked when other faults are active or expose itself only later~\citep{xu2016configuration}.

\section{The alert rule pack}
\label{app:alerts}

The frozen pack holds \val{alerts.total} rules built from the \val{alerts.templates} templates of Table~\ref{tab:alerts}: \val{alerts.templates.shared} apply to every application and \val{alerts.templates.astro_only} to Astronomy Shop only, whose Kafka and rollout metrics the other two applications do not expose. Each rule is scoped to its application's namespaces, the request rules read the application's entry-point service, and ``holds for'' is how long a condition must hold before the alert fires.

\begin{table}[!htbp]
\caption{The rule templates of the frozen pack. The condition paraphrases the PromQL expression, and the error-share rules also require a minimum request rate.}
\label{tab:alerts}
\begin{center}
\scriptsize
\setlength{\tabcolsep}{4pt}
\begin{tabular}{@{}l>{\raggedright\arraybackslash}p{0.39\linewidth}lll@{}}
\toprule
Rule & Fires when & Severity & Holds for & Applications \\
\midrule
\texttt{HighRequestErrorRate} & 5xx share of entry-point requests above 10\% & Critical & 1m & All \\
\texttt{EndpointErrorRate} & 5xx share of one entry-point route above 50\% & Critical & 1m & All \\
\texttt{HighRequestLatency} & 95th-percentile latency of a service above 3000\,ms & Critical & 1m & All \\
\texttt{PodStatusError} & A container is waiting with a reported reason & Critical & 1m & All \\
\texttt{FailedPodsDetected} & A pod is in phase Failed & Critical & 1m & All \\
\texttt{DeploymentNotReady} & A deployment with replicas requested has none available & Critical & 1m & All \\
\texttt{EndpointErrorRateSensitive} & 5xx share of one entry-point route above 10\% & Warning & 1m & All \\
\texttt{HighRequestRate} & Entry-point request rate above 50 per second & Warning & 1m & All \\
\texttt{ServiceLatencyElevated} & 90th-percentile latency of a service above 1000\,ms & Warning & 1m & All \\
\texttt{PendingPodsDetected} & A pod is in phase Pending & Warning & 1m & All \\
\texttt{PodSchedulingFailure} & A pod cannot be scheduled & Warning & 30s & All \\
\texttt{PodContainerRestarting} & A container restarted in the last 10 minutes & Warning & 1m & All \\
\texttt{PodNotReady} & A pod is not ready & Warning & 2m & All \\
\midrule
\texttt{AstronomyShopTelemetryDown} & A scrape target of the telemetry stack is down & Critical & 30s & Astro \\
\texttt{ContainerOOMKilled} & A container restarted after being killed for exceeding its memory limit & Critical & 1m & Astro \\
\texttt{MessageConsumerLag} & Kafka consumer lag above 1000 records on a partition & Warning & 2m & Astro \\
\texttt{DeploymentGenerationMismatch} & A deployment's latest revision has not been observed by its controller & Warning & 5m & Astro \\
\texttt{DeploymentReplicasUnavailable} & A deployment has fewer available replicas than requested & Warning & 5m & Astro \\
\bottomrule
\end{tabular}

\end{center}
\end{table}

\section{Instrumented tool details}
\label{app:tools}

Table~\ref{tab:tools} lists the \val{tools.published} tools the benchmark publishes, generated from the tool registry. Read tools are free and logged for cost accounting only. The report tool writes to the scored record and cannot touch the system, so it counts toward neither Intervention risk nor the per-invocation action budget. Act and guarded calls are logged and count toward Intervention risk (\S\ref{sec:evaluation}), and every guarded call must cite an incident the agent has already reported. The forbidden operations are \val{tools.forbidden}, and any attempt at the three counts as a violation. Anything the agent writes to its workspace but does not report through \texttt{incident\_report} does not exist for scoring.

\begin{table}[!htbp]
\caption{The published tools by tier. The policy enforcement layer computes every mutating call's target set from its typed arguments, and every guarded call must name a reported incident.}
\label{tab:tools}
\begin{center}
\scriptsize
\setlength{\tabcolsep}{4pt}
\begin{tabular}{@{}llp{0.62\linewidth}@{}}
\toprule
Tool & Tier & Function \\
\midrule
\texttt{code\_search} & Read & Search the source of one service. \\
\texttt{get\_alerts} & Read & Alerts fired in full cumulative alert history, with their labels. \\
\texttt{get\_topology} & Read & The service call graph derived from traces. \\
\texttt{kubectl\_describe} & Read & Describe one Kubernetes object with its recent events. \\
\texttt{kubectl\_get} & Read & Read the current state of one Kubernetes object. \\
\texttt{query\_logs} & Read & Recent log lines of one deployment. \\
\texttt{query\_metrics} & Read & Evaluate an instant PromQL query. \\
\texttt{query\_traces} & Read & Recent traces of one service. \\
\texttt{read\_file} & Read & Read one file of a service's source. \\
\texttt{search\_logs} & Read & Evaluate a LogQL query across services. \\
\texttt{incident\_report} & Report & Publish the scored record: an alert-to-incident mapping, a root-cause claim, or an incident status. \\
\texttt{restart\_pod} & Act & Restart the pods of one deployment. \\
\texttt{build\_service\_image} & Guarded & Build a service image that includes a proposed patch, without deploying it. \\
\texttt{propose\_patch} & Guarded & Apply a diff to an isolated clone of one service's source. \\
\texttt{rollback} & Guarded & Return one deployment to its previous revision. \\
\texttt{scale\_deployment} & Guarded & Set the replica count of one deployment. \\
\texttt{set\_feature\_flag} & Guarded & Set one feature flag and restart the services that read it. \\
\texttt{submit\_patch} & Guarded & Apply a diff to one service, then build and deploy its image through the release pipeline. \\
\texttt{update\_config} & Guarded & Change one field of a Deployment, Service, or ConfigMap. \\
\bottomrule
\end{tabular}

\end{center}
\end{table}

\FloatBarrier
\section{Worked examples for every scored formula}
\label{app:worked}

We work every quantity of \S\ref{sec:evaluation} on one small run: the orchestrator injects four faults, $f_1$ to $f_4$, and the agent's final report holds three incidents, $I_1$ to $I_3$.

\paragraph{Correlated.} $I_1$ holds an alert attributable to $f_1$, so $a_1=1$. $I_2$ holds alerts attributable to both $f_2$ and $f_3$. Faults and incidents are matched one-to-one, so only one of the two is matched, here $f_2$: $a_2=1$ and $a_3=0$. $I_3$ holds only an alert attributable to no fault, and no incident holds an alert of $f_4$, so $a_4=0$. $\mathrm{Corr}=2/4=0.500$. Filing every alert under one incident would have matched one fault at most.

\paragraph{Located.} The diagnosis associated with $f_1$ names its culprit service and its responsible artifact, so $b_1=1$. The diagnosis associated with $f_2$ names the right service and the wrong artifact, so $b_2=0$. No diagnosis names $f_3$ or $f_4$, so $b_3=b_4=0$. $\mathrm{Loc}=1/4=0.250$.

\paragraph{Repaired.} The agent rolls back the culprit service of $f_1$, and the fault's symptom check returns to healthy and remains healthy until the orchestrator reverts the fault: $c_1=1$. It restarts the service of $f_2$, and the symptom check returns to healthy but fails again before the revert: $c_2=0$. The symptom check of $f_3$ recovers only when the orchestrator reverts the fault, and benchmark cleanup is never credited: $c_3=0$. The symptom check of $f_4$ never failed, and its responsible artifact was not verified to have been corrected: $c_4=0$. $\mathrm{Rep}=1/4=0.250$.

\paragraph{\ourScore.} $f_1$ passes all three stages, so it is handled and earns $\tfrac{1+1+1}{3}=1$. $f_2$ is correlated but not located, so it earns $\tfrac{1}{3}$. $f_3$ and $f_4$ are not correlated and earn nothing, as would a fault located without being correlated. The score is $100\cdot\tfrac{1}{4}\cdot\bigl(1+\tfrac{1}{3}\bigr)=33.3$. Had the run attempted one forbidden operation, blocked or not, $d=1$ and the score would be $100\cdot\tfrac{1}{2}\cdot\tfrac{1}{4}\cdot\bigl(1+\tfrac{1}{3}\bigr)=16.7$.

\paragraph{Intervention risk.} The agent executed ten interventions, two outside the affected scope of the incident they referenced, and attempted one forbidden operation: $n_{\mathrm{exec}}=10$, $n_{\mathrm{out}}=2$, $n_{\mathrm{forb}}=1$, and $\mathrm{Risk}=(2+1)/(10+1)=0.273$.

\paragraph{Service degradation.} The integral is taken by the trapezoidal rule over the recorded timestamps. Suppose the entry-point service serves $q=10$ requests/s and $e+\ell$ is 0, 0.2, and 0 at 0, 60, and 120\,s. Then $\min\{1,e+\ell\}\,q$ is 0, 2, and 0 requests/s, and $D_{\mathrm{agent}}=60\cdot\tfrac{0+2}{2}+60\cdot\tfrac{2+0}{2}=120$ degraded requests. Against $D_{\mathrm{no\text{-}action}}=240$ from the no-action run with the same application configuration and fault-schedule seed, the degradation level is $120/240=50\%$. With $D_{\mathrm{agent}}=300$ it would be $300/240=125\%$: the agent left users worse off than no agent would have.

\paragraph{MTTD, MTTL, and MTTR.} Each median is taken over the faults that reached the corresponding stage, so MTTD counts only correlated faults. The first incident reports containing an alert of $f_1$ and of $f_2$ come 120 and 360\,s after their injections. The alert of $f_3$ also sits in $I_2$, but $f_3$ is not correlated and has no detection time. The median is the $\lceil 0.5\,n\rceil$-th smallest time, so it is always an observed time: with $n=2$ it is the first, and $\mathrm{MTTD}=120$\,s over two faults. $f_1$ is the only fault located and the only one repaired, at 240 and 480\,s, so $\mathrm{MTTL}=240$\,s and $\mathrm{MTTR}=480$\,s, each over one fault. A stage that no fault passed has no time and prints a dash.

\paragraph{Cost.} The run made 200 model calls with 5.0M input and 0.1M output tokens, priced at \$3.10, so it reports 200 calls, 5.1M tokens, and \$3.10; Table~\ref{tab:pooled} prints the tokens and dollars.

\section{Baselines and coding agents}
\label{sec:orbit}

\paragraph{Runbook.} Runbook is inspired by SRE day-to-day workflow: it calls no model and groups alerts into one incident per workload their labels name, and opens a new incident when a closed one's workload alerts again. A fixed table (Table~\ref{tab:runbook}) maps each alert rule to one scripted action and a blamed fault category: roll back on error, latency, and stalled-rollout alerts, scale up on unavailable-deployment and load alerts, and restart on pod failures and any other alert. When several rules fire, a rule naming the broken object wins over a symptom rule. For each new incident it reports the workload and the table's category as the root cause, runs the action once, and marks the incident resolved at the next invocation if the action succeeded.

\begin{table}[!htbp]
\caption{Runbook's rule table in priority order: when an incident's alerts fire several rules, the first listed wins. The Online Boutique and Train Ticket rules carry an application prefix and take the row of their unprefixed name.}
\label{tab:runbook}
\begin{center}
\scriptsize
\setlength{\tabcolsep}{4pt}
\begin{tabular}{@{}lll@{}}
\toprule
Alert rule & Action & Reported category \\
\midrule
\texttt{DeploymentNotReady} & \texttt{scale\_deployment} to 2 & \texttt{deployment} \\
\texttt{DeploymentReplicasUnavailable} & \texttt{scale\_deployment} to 2 & \texttt{deployment} \\
\texttt{DeploymentGenerationMismatch} & \texttt{rollback} & \texttt{deployment} \\
\texttt{ContainerOOMKilled} & \texttt{restart\_pod} & \texttt{code-bug} \\
\texttt{PodStatusError} & \texttt{restart\_pod} & \texttt{code-bug} \\
\texttt{PodNotReady} & \texttt{restart\_pod} & \texttt{code-bug} \\
\texttt{FailedPodsDetected} & \texttt{restart\_pod} & \texttt{code-bug} \\
\texttt{PodContainerRestarting} & \texttt{restart\_pod} & \texttt{code-bug} \\
\texttt{PodSchedulingFailure} & \texttt{restart\_pod} & \texttt{infrastructure} \\
\texttt{PendingPodsDetected} & \texttt{scale\_deployment} to 2 & \texttt{infrastructure} \\
\texttt{MessageConsumerLag} & \texttt{scale\_deployment} to 2 & \texttt{infrastructure} \\
\texttt{AstronomyShopTelemetryDown} & \texttt{restart\_pod} & \texttt{infrastructure} \\
\texttt{HighRequestErrorRate} & \texttt{rollback} & \texttt{code-bug} \\
\texttt{EndpointErrorRate} & \texttt{rollback} & \texttt{code-bug} \\
\texttt{EndpointErrorRateSensitive} & \texttt{rollback} & \texttt{code-bug} \\
\texttt{HighRequestLatency} & \texttt{rollback} & \texttt{configuration} \\
\texttt{ServiceLatencyElevated} & \texttt{rollback} & \texttt{configuration} \\
\texttt{HighRequestRate} & \texttt{scale\_deployment} to 2 & \texttt{infrastructure} \\
\midrule
\textit{any other alert} & \texttt{restart\_pod} & \texttt{infrastructure} \\
\bottomrule
\end{tabular}

\end{center}
\end{table}

\paragraph{ReAct.} ReAct~\citep{yao2023react} starts a fresh conversation at every invocation. Its first message holds the cumulative alert history and the incidents it opened earlier, which are the only state it keeps between invocations. Before the model runs, new alerts are filed under the first unresolved incident, so every alert has an incident even if the model reports none. The model then alternates reasoning and tool calls over the full tool set of Appendix~\ref{app:tools}, choosing every call itself, until it stops calling tools, ten steps pass, or 150k tokens are spent. It reports incidents, root causes, and statuses through the report tool.

\paragraph{RAG.} RAG~\citep{lewis2020rag} makes one model call per invocation, and the model calls no tools. It retrieves from the live telemetry of that invocation and reads the topology, then the logs and traces of up to five services named in the alerts, most recently alerting first, within 45\,s. Each tool answer, cut to 1200 characters, is one document. Then it ranks the documents against a query made of every alert's rule name and label values, and the top five enter the prompt beside the alerts and the incidents of earlier invocations. The model returns one plan that groups alerts into incidents and gives each a root cause, an optional repair from the act and guarded tools, and a status. The code publishes and executes the plan as written.

\paragraph{Keep-agent.} Keep~\citep{keephq2026keep} is an open-source alert-management platform: a rule engine groups alerts into incidents, and workflows run scripted actions. Its model-based correlation is a hosted service outside the open-source release, so we do not evaluate it. Keep-agent runs Keep inside the agent's process at every invocation, keeps its incident store in the persistent workspace, and sends it each alert once. The rules are ours, written in Keep's rule language: one rule per label that names an object (the workload labels first, then the pod, then the request path), each excluding the labels ranked above it, so an alert joins at most one incident. An incident stays open for the whole run, so a second fault on the same workload joins the first fault's incident, and alerts that name no object wait in an unassigned bucket. To this we add one model call per invocation with RAG's prompt, telemetry retrieval, and plan format, unchanged. The prompt lists Keep's incidents as fixed: the model may place only the alerts no rule grouped, and the code discards any move of an alert Keep placed. The model gives each incident a root cause, an optional repair, and a status. Each repair is sent through a Keep workflow whose one HTTP step calls the agent--system interface, so the policy enforcement layer checks it as it checks every method's. Keep-agent and RAG therefore share the model's half and differ in who groups the alerts.

\begin{figure}[!htb]
\centering
\resizebox{\linewidth}{!}{%
\begin{tikzpicture}[
  font=\scriptsize,
  stage/.style={draw, rounded corners=2pt, align=center, inner sep=3pt, minimum height=9mm, text width=24mm, fill=white},
  store/.style={draw, dashed, rounded corners=2pt, align=center, inner sep=3pt, minimum height=8mm, fill=black!4},
  arr/.style={-{Stealth[length=1.6mm]}, semithick},
  lbl/.style={font=\tiny, text=black!60},
  node distance=5mm
]
\node[stage] (merge) {\textbf{Merge alerts}\\ one incident per signal};
\node[stage, right=of merge] (floor) {\textbf{Publish mapping}\\ before any model call};
\node[stage, right=of floor] (loop) {\textbf{Decision loop}\\ $\le$10 typed JSON steps};
\node[stage, right=of loop] (yield) {\textbf{Publish and learn}\\ status per incident, lessons};
\node[store, text width=44mm, below=7mm of $(merge.south)!0.5!(floor.south)$] (ws) {\textbf{Workspace} (survives the per-invocation restart)\\ agenda $\cdot$ incidents $\cdot$ ledger $\cdot$ memory $\cdot$ charter};
\node[store, text width=24mm, below=7mm of loop.south, xshift=5mm] (tools) {\textbf{Tools}\\ read $\cdot$ act $\cdot$ guarded\\ one bounded subagent};
\draw[arr] ([xshift=-9mm]merge.west) -- (merge.west) node[lbl, pos=0, above right, xshift=-9mm] {alert batch at $t$};
\draw[arr] (merge) -- (floor);
\draw[arr] (floor) -- (loop);
\draw[arr] (loop) -- (yield);
\draw[arr] (yield.east) -- ([xshift=6mm]yield.east) node[lbl, pos=1, right] {yield};
\draw[arr] (ws.north -| merge.south) -- (merge.south) node[lbl, midway, left] {load};
\draw[arr, <->] (loop.south west) |- (ws.east) node[lbl, pos=0.22, right] {state};
\draw[arr, <->] (loop.south -| tools.north) -- (tools.north) node[lbl, midway, right] {calls};
\draw[arr] (yield.south) |- ($(ws.south)+(0,-3mm)$) node[lbl, pos=0.6, below] {save} -- (ws.south);
\end{tikzpicture}}
\vspace{-2mm}
\caption{One invocation of \textsc{Orbit}. Alert merging and both publication steps are deterministic, and the model decides only inside the loop. Everything kept across invocations lives in the workspace.}
\label{fig:orbit}
\end{figure}
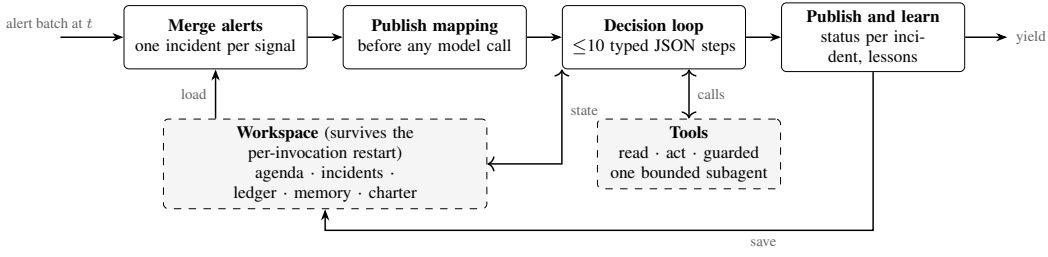

\paragraph{\textsc{Orbit}.} We built \textsc{Orbit} as an agent that covers every stage the benchmark scores, since no existing agent is designed to correlate, locate, and repair across invocations of a continuous run. Unlike ReAct, which starts a fresh conversation at every invocation, it keeps state across invocations and organizes its own work. Figure~\ref{fig:orbit} shows one invocation. The benchmark starts a fresh process per invocation, and \textsc{Orbit} reloads a checkpoint, an agenda, its open incidents, and a durable action ledger from its workspace. An action that succeeded while the previous process died is folded back into its incident first. The alert batch is merged deterministically, one incident and one agenda task per signal, and the complete alert-to-incident mapping is published before the model is consulted, so what Correlated grades never depends on the model finishing.

The model then works inside a bounded decision loop. Each step sends a fixed operating charter, the state, the last observations, the tool catalog, and the remaining budget, and expects one JSON decision: a \texttt{tool\_call} naming an agenda task, the incident it serves, the expected effect, and the check that will confirm it, or a \texttt{finish} returning the rewritten agenda, mapping, and claims. A malformed decision gets one correction call, and a second failure ends the invocation as a charged no-op. Ten steps, a token ceiling, and a mutation cap bound the loop, and a budget ladder reduces it to triage only as the budget runs out. Source-level fixes happen only in one subagent bound to a diagnosed incident and service. Before yielding, \textsc{Orbit} publishes a status and at least a tentative root cause per open incident, appends at most three lessons to a bounded charter, and promotes a repair recipe to procedural memory only once the incident it resolved has validated green. The contrast with ReAct is the point: same tools and model, but state, protocol, and memory instead of a fresh conversation every invocation.

\paragraph{Codex.} We run release \texttt{rust-v0.154.0} of Codex~\citep{openai2026codex} unmodified in headless mode, behind a thin adapter that configures, prompts, and relays. At each invocation the benchmark starts a fresh Docker container on its host, outside the Kubernetes cluster, from an image holding the Codex binary and the adapter; the container sees only the run's workspace and the variables the benchmark passes, and is stopped at the invocation's time limit. Its only tools are those of Appendix~\ref{app:tools}, reached through an MCP server the adapter starts; its shell and web search are disabled, so every call passes the policy enforcement layer. The benchmark's rules are given as project instructions, and one Codex conversation, stored in the workspace, is resumed at every invocation, so its own context is its memory across incidents. Each invocation must end with a JSON block listing its incidents, root causes, and statuses, which the adapter publishes through the report tool, and a repair is held until the incident it cites has been reported. Model calls go through the metering proxy, and an invocation stops at 150k tokens or 40 tool calls.

\paragraph{Claude Code.} We run release 2.1.273 of Claude Code~\citep{anthropic2026claudecode} behind the same adapter as Codex, with the same tools, report block, repair ordering, and limits, counting 40 model calls rather than tool calls. It differs from Codex in its memory and local tools. Each invocation starts a fresh session, and the agent's memory is a notes directory it writes itself. It runs in its restricted mode, with only its file tools (read, write, edit, and search) confined to that directory and no shell or web tools. A local shim forwards its API calls to the metering proxy and returns the answers in the format Claude Code expects.

\section{How far a single run generalizes}
\label{app:robustness}

Table~\ref{tab:react-seeds} gives ReAct on Astronomy Shop under the shared seed of Table~\ref{tab:pooled} and two further seeds. Because the fault catalog is large, the schedules that different seeds draw differ widely, and so do the results: the stage shares differ by a factor of two to four between seeds, Intervention risk by more than four, and the \ourScore{} from \val{ms.17.react-b2.kimi.Astro.score} to \val{ms.7.react-b2.kimi.Astro.score}. This is why every method is compared under one shared schedule, and it is also what makes the benchmark usable for reinforcement learning, since each new seed is a rollout under a different setting.

\begin{table}[!htbp]
\caption{ReAct on Kimi K2.5, one Astronomy Shop marathon run per seed. Columns are as in Table~\ref{tab:pooled}.}
\label{tab:react-seeds}
\begin{center}
\footnotesize
\setlength{\tabcolsep}{4.2pt}
\begin{tabular}{@{}llrrrrrrrr@{}}
\toprule
Seed & Venue & Corr & Loc & Rep & MTTD & MTTL & MTTR & Risk & \ourScore \\
\midrule
Shared & Single-host & 0.517 & 0.121 & 0.017 & 231 & 329 & 709 & 0.136 & 20.1 \\
17 & Single-host & 0.203 & 0.034 & 0.051 & 202 & 625 & 593 & 0.593 & 7.3 \\
27 & Multi-node & 0.455 & 0.073 & 0.018 & 208 & 324 & 2486 & 0.250 & 17.0 \\
\bottomrule
\end{tabular}

\end{center}
\end{table}

\section{Agent behavior under lighter pressure}
\label{app:short}

We use short runs to see how an agent behaves under lighter pressure than a marathon run. A short run injects five or ten faults, one per invocation, and holds every one open to the end of the run, so the agent has the rest of the run to repair each fault and no fault clears on its own. Each short run is paired with a no-action run on the same schedule and seed that supplies its degradation reference. Six short runs hold five faults and six hold ten, two per application (Table~\ref{tab:storm-drill}).

\begin{table}[!htbp]
\caption{ReAct on Kimi K2.5 by schedule regime. A short-run row averages six short runs, two per application, and the marathon-run row averages the three marathon runs of Table~\ref{tab:pooled}. Columns are as in Table~\ref{tab:pooled}.}
\label{tab:storm-drill}
\label{tab:short-pooled}
\begin{center}
\small
\begin{tabular}{@{}lrrrrr@{}}
\toprule
Regime & Runs & Corr & Loc & Rep & \ourScore \\
\midrule
Short run, 5 faults & 6 & 0.800 & 0.200 & 0.500 & 38.9 \\
Short run, 10 faults & 6 & 0.728 & 0.283 & 0.183 & 34.8 \\
Marathon run, about 60 faults & 3 & 0.694 & 0.079 & 0.006 & 25.4 \\
\bottomrule
\end{tabular}

\end{center}
\end{table}

Under lighter pressure the same ReAct agent repairs far more and reaches a higher \ourScore. With five faults it repairs a share of \val{ms.held5.react.repaired} and scores \val{ms.held5.react.score}, with ten faults \val{ms.held10.react.repaired} and \val{ms.held10.react.score}, and in the marathon runs \val{ms.storm.react.repaired} and \val{ms.storm.react.score}. The share of faults handled end to end falls from \val{ms.held5.react.handled}\% to \val{ms.held10.react.handled}\% and \val{ms.storm.react.handled}\%. Correlated moves far less across the three regimes (\val{ms.held5.react.correlated}, \val{ms.held10.react.correlated}, and \val{ms.storm.react.correlated}), so what pressure removes is repair rather than the grouping of alerts. The decline is already visible with ten faults, where the commonest outcome is an executed action that does not repair (\val{rq.react-b2.held10.outcome.acted.pct}\% of faults, Appendix~\ref{app:decomposition}). The short runs differ from the marathon runs in venue, release, seed, invocation interval, and fault lifetime as well as in load, so we read this as a qualitative contrast rather than a controlled measurement of concurrency.

\section{Where the lifecycle breaks}
\label{app:decomposition}

\begin{figure}[!htb]
\centering
\includegraphics[width=\linewidth]{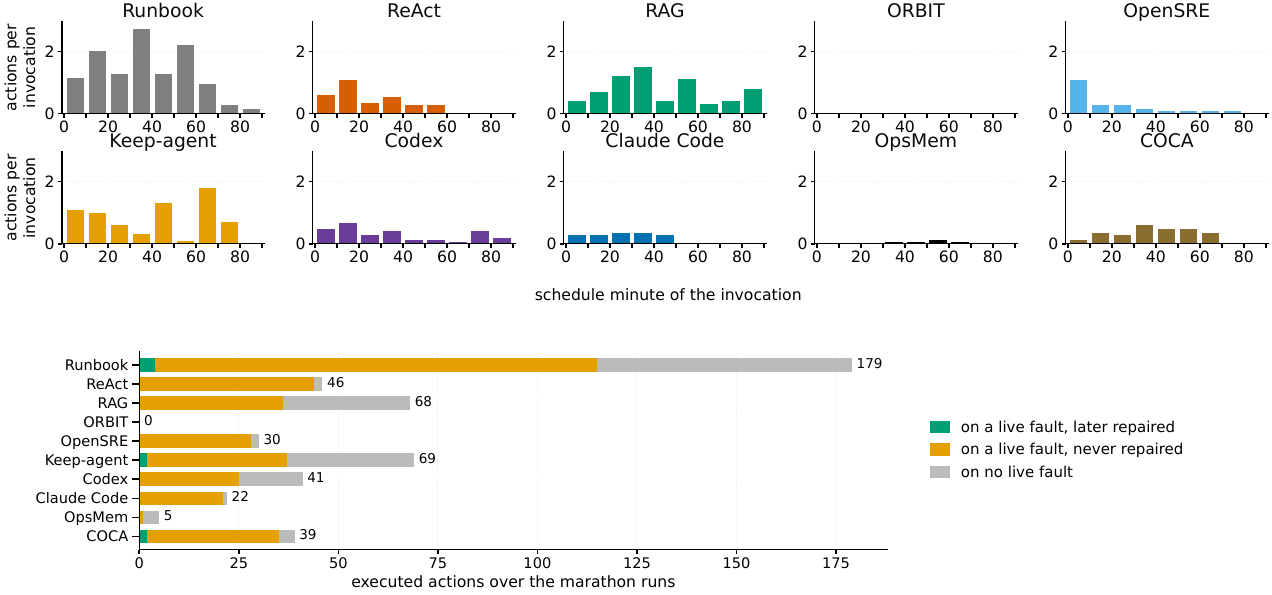}
\vspace{-4mm}
\caption{
Every method of Table~\ref{tab:pooled} on Kimi K2.5, and Runbook. Top: executed actions per invocation in the ten-minute schedule bins of Figure~\ref{fig:activity-time}; unlike its act tier, calls that were denied or failed are not counted. Bottom: every executed
action of a method's marathon runs by what it touched, with the total at the bar's end. An action
is \emph{on a live fault} when its target set holds the faulted service of a fault injected and not
yet reverted. Matching is by service, so it says where an action went and not whether the remedy fit.
}
\label{fig:repair-effect}
\end{figure}

A repaired share near zero, as in every marathon run, can mean that agents stop acting or that they keep acting to no effect. Figure~\ref{fig:repair-effect} separates the two for the runs of \S\ref{sec:results-time}. Its top row counts executed actions per invocation, leaving out denied or failed calls: in minutes 60--70, Keep-agent makes \val{rq.keep-llm.kimi.60-70.act} system-changing calls per invocation in Figure~\ref{fig:activity-time} but executes \val{rq.keep-llm.kimi.60-70.executed}. Its bottom panel sorts every executed action by what it touched.

\paragraph{Agents act on the right service and still repair almost nothing.} \textsc{Orbit} never acts: on Kimi K2.5 its model spends each invocation's few decisions on reads, reports, and subagent hand-offs, and on \gptluna{} it hands repairs to a source-fix subagent that never submits a patch. OpsMem executes only \val{rq.opsmem.kimi.actions} actions in three runs. Every other method acts through most of the injection period, and more than half of its actions hit the faulted service of a live fault; \val{rq.react-b2.kimi.actions.unrepaired} of ReAct's \val{rq.react-b2.kimi.actions} actions and \val{rq.opensre-x3.kimi.actions.unrepaired} of OpenSRE's \val{rq.opensre-x3.kimi.actions} land on faults that are never repaired. Runbook acts most (\val{rq.b1-runbook.none.actions} actions), followed by Keep-agent (\val{rq.keep-llm.kimi.actions}) and RAG (\val{rq.rag-b2r.kimi.actions}), though \val{rq.keep-llm.kimi.actions.off.pct}\% and \val{rq.rag-b2r.kimi.actions.off.pct}\% of the latter two's actions hit no live fault. Only \val{rq.b1-runbook.none.actions.repaired} of Runbook's actions reach a fault that is later repaired; among the LLM-based methods, only Keep-agent and COCA on Kimi K2.5 (\val{rq.keep-llm.kimi.actions.repaired} and \val{rq.coca.kimi.actions.repaired}) and Keep-agent and Codex on \gptluna{} (\val{rq.keep-llm.luna.actions.repaired} and \val{rq.codex-x4.luna.actions.repaired}, not shown in Figure~\ref{fig:repair-effect}) have any such action. Every other repaired fault in Table~\ref{tab:pooled} was credited through an action on another service in its affected scope, which matching by faulted service misses. When these methods act, they usually reach the right service but apply a remedy that does not fix it.

\begin{figure}[!htb]
    \centering
    \includegraphics[width=0.72\linewidth]{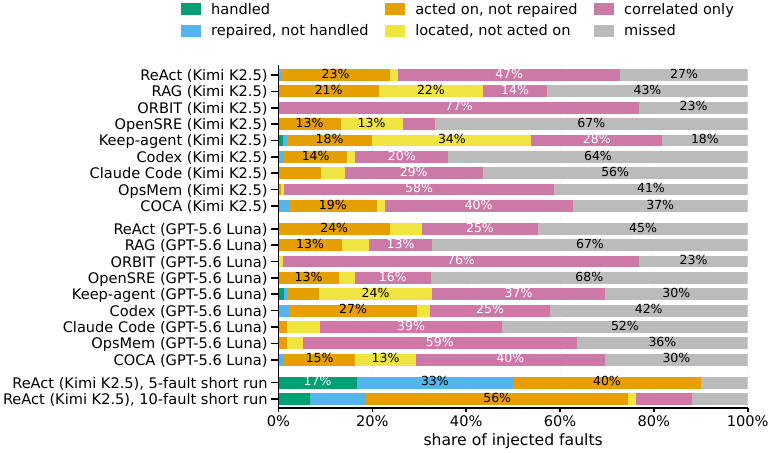}
    \caption{One outcome per injected fault, by the first rule that matches: handled, meaning all three stages passed; repaired but not handled; acted on, meaning an executed action targeted its faulted service while it was live, but not repaired; located but not acted on; correlated only; or missed. Rows are the LLM-based methods of Table~\ref{tab:pooled} on both backbones, plus the two ReAct short-run rows.}
    \label{fig:outcomes}
\end{figure}

\paragraph{Methods with similar scores stall at different stages.} Figure~\ref{fig:outcomes} assigns each injected fault one outcome. On Kimi K2.5, ReAct stops at correlation: \val{rq.react-b2.kimi.outcome.correlated.pct}\% of its faults are grouped and taken no further, and \val{rq.react-b2.kimi.outcome.acted.pct}\% are acted on without a repair. RAG, OpenSRE, Codex, and Claude Code most often miss the fault (\val{rq.rag-b2r.kimi.outcome.missed.pct}\%, \val{rq.opensre-x3.kimi.outcome.missed.pct}\%, \val{rq.codex-x4.kimi.outcome.missed.pct}\%, and \val{rq.claudecode-x5.kimi.outcome.missed.pct}\%), while OpsMem and COCA most often stop at correlation (\val{rq.opsmem.kimi.outcome.correlated.pct}\% and \val{rq.coca.kimi.outcome.correlated.pct}\%). Keep-agent misses the fewest (\val{rq.keep-llm.kimi.outcome.missed.pct}\%) and has the largest share of faults it locates without acting on them (\val{rq.keep-llm.kimi.outcome.located.pct}\%). \textsc{Orbit} correlates \val{rq.orbit-0.kimi.outcome.correlated.pct}\% of its faults and does nothing further. On \gptluna{}, seven of the nine methods keep their most common outcome. The exceptions are ReAct, which misses more faults (\val{rq.react-b2.luna.outcome.missed.pct}\%) than it leaves at correlation (\val{rq.react-b2.luna.outcome.correlated.pct}\%), and Keep-agent, which leaves more at correlation (\val{rq.keep-llm.luna.outcome.correlated.pct}\%) than it locates without acting (\val{rq.keep-llm.luna.outcome.located.pct}\%). In the two short-run rows, the same ReAct agent handles \val{rq.react-b2.held5.outcome.handled.pct}\% of faults with five held and \val{rq.react-b2.held10.outcome.handled.pct}\% with ten (Appendix~\ref{app:short}). Similar scores therefore hide different failures: some methods group alerts and stop, several never find the fault, Keep-agent locates many faults it does not act on, and the methods that act apply remedies that do not repair.

\section{Limitations}
\label{sec:limitations}

\ourBench{} makes several compromises that bound what its results mean. We group them by where they enter: the environment, the faults, the alerting, the scoring, and the methods we ran.

\paragraph{Environment.} On the single-host venue the two zones share one machine, so a zone failure is an emulation: the redundancy semantics are real even where the physics are not. The three applications are public, so memorization is a live risk that newly written code faults and exact-commit artifact targets reduce but do not remove.

\paragraph{Faults.} The category mix follows one provider's high-severity incidents over 2020--21, the best public anchor we know of, but only one. A fault's affected scope is measured one fault at a time and is not re-measured under overlap, so attribution is conservative when faults interact.

\paragraph{Alerting.} The frozen rule pack sets the difficulty of correlation: a fault below every rule's threshold raises no alert, as a pilot fault showed, and the pack is versioned with the release for that reason. The pack is fixed for the agent as well, since no tool lets it add a rule. Letting a long-horizon agent extend monitoring coverage as it learns the system, with its own alerts kept out of the scored denominator, is a promising direction that the design anticipates but the benchmark does not yet score.

\paragraph{Scoring.} Located matches a diagnosis exactly against the release's alias table, so a diagnosis that names the right cause in other words scores zero. We chose exactness over a model judge so that every leaderboard number rests on recorded evidence alone. Repaired credits a rollback and a corrected patch alike. The degradation level is measured against one no-action recording per release.

\paragraph{Backbones and baselines.} We chose to run the agents on two LLM backbones, Kimi K2.5 and \gptluna{}, and neither is the most capable model available. We prioritized budget-friendly models over frontier ones in consideration of cost-effectiveness in real-world scenarios where SREs deal with large volumes of alerts per day. The ten methods span operations practice, reasoning patterns, specialized SRE agents, coding agents, and diagnosis methods, but they cannot cover the entire field, and a method built for this setting may well do better than any reported here.

\end{document}